\documentclass[runningheads]{llncs}

\usepackage{graphicx}
\usepackage{amsmath}
\usepackage{array}
\usepackage{multirow}
\usepackage{caption}
\usepackage{subcaption}
\usepackage{xcolor}
\usepackage{url}
\usepackage{booktabs}
\newcommand{\plus}{\raisebox{.4\height}{\scalebox{.6}{+}}}

\begin{document}

\title{Individualized multi-horizon MRI trajectory prediction for Alzheimer's Disease}

\titlerunning{Individualized multi-horizon MRI prediction}

\author{Rosemary He\inst{1,2}\orcidID{0000-0001-8307-3958}
\and
Gabriella Ang\inst{1}\orcidID{0009-0009-7901-6809}
\and
Daniel Tward\inst{2,3,4}\orcidID{0000-0002-4607-6807}
\and
for the Alzheimer's Disease Neuroimaging Initiative\thanks{Data used in preparation of this article were obtained from the Alzheimer's Disease Neuroimaging Initiative(ADNI) database \url{adni.loni.usc.edu}). As such, the investigators within the ADNI contributed to the design and implementation of ADNI or provided data but did not participate in analysis or writing of this report. A complete listing of ADNI investigators can be found at: \url{http://adni.loni.usc.edu/wp-content/uploads/how\_to\_apply/ADNI\_Acknowledgement\_List.pdf}}}

\authorrunning{R. He et al.}

\institute{Departments of Computer Science, University of California, Los Angeles \and
Departments of Computational Medicine, University of California, Los Angeles \and
Departments of Neurology, University of California, Los Angeles \and
Corresponding author, DTward@mednet.ucla.edu}

\maketitle

\begin{abstract}
Neurodegeneration as measured through magnetic resonance imaging (MRI) is recognized as a potential biomarker for diagnosing Alzheimer’s disease (AD), but is generally considered less specific than amyloid or tau based biomarkers. Due to a large amount of variability in brain anatomy between different individuals, we hypothesize that leveraging MRI time series  can help improve specificity, by treating each patient as their own baseline. Here we turn to conditional variational autoencoders to generate individualized MRI predictions given the subject’s age, disease status and one previous scan. Using serial imaging data from the Alzheimer’s Disease Neuroimaging Initiative*, we train a novel architecture to build a latent space distribution which can be sampled from to generate future predictions of changing anatomy. This enables us to extrapolate beyond the dataset and predict MRIs up to 10 years. We evaluated the model on a held-out set from ADNI and an independent dataset (from Open Access Series of Imaging Studies). 
By comparing to several alternatives, we show that our model produces more individualized images with higher resolution. Further, if an individual already has a follow-up MRI, we demonstrate a usage of our model to compute a likelihood ratio classifier for disease status. In practice, the model may be able to assist in early diagnosis of AD and provide a counterfactual baseline trajectory for treatment effect estimation. Furthermore, it generates a synthetic dataset that can potentially be used for downstream tasks such as anomaly detection and classification.

\keywords{Conditional VAE \and Disease progression \and Alzheimer's Disease.}
\end{abstract}

\section{Introduction}

\subsection{Background}
Neurodegeneration as observed in magnetic resonance imaging (MRI) is recognized as a potential biomarker for diagnosing Alzheimer's disease (AD)\cite{jack2011introduction}. While MRI has the advantage of being noninvasive, it is generally considered not specific enough, unlike biomarkers of amyloid or tau\cite{jack2018nia}. The potential for using machine learning approaches (``data-driven statistical approaches in which many different brain regions are evaluated simultaneously''\cite{albert2011diagnosis}) has been recognized. However, an important challenge in making diagnoses from brain images is the large amount of interpersonal variability, compared to minor structural changes in the earliest stages of the disease such as the transentorhinal stage (Braak stage I and II)\cite{braak1991neuropathological}. We hypothesize that this issue can be overcome by carefully modeling timeseries of imaging data, allowing each person to be compared to their own baseline, rather than to a population average.

\subsection{Related work} \label{relate_work}
One strategy for timeseries analysis in brain imaging is to quantify structural changes using the diffeomorphism group\cite{beg2005computing}. After defining an appropriate metric, geodesics can be computed to study changes over time\cite{thomas2013geodesic,durrleman2013toward,tward2017entorhinal}. More recently, deep learning has been applied to model disease progression via changing pixel values, with models including deep structural causal models\cite{abdulaal2022deep}, variational autoencoders (VAE)\cite{zhao2019variational,sauty2022progression}, generative adversarial networks (GAN)\cite{xia2019consistent,ravi2019degenerative}, and diffusion models\cite{yoon2023sadm,puglisi2024enhancing}. Among these works, we highlight two that train on time-series MRIs to predict the aging brain, conditioned on age and diagnosis. The first \cite{sauty2022progression} combines a variational autoencoder with a mixed-effect model imposed on the latent space to simulate the aging process. By explicitly modeling age in the latent space, this model synthesizes images across 30+ years with atrophy patterns consistent with expectation. 
However, the model focuses on providing a population trajectory and is not specific enough for individual predictions, and images produced tend to be lower resolution as is typical of VAEs. In the second, 4D-DANI-Net, a GAN model with specified loss functions and a weight profile function, is proposed to simulate the aging process both globally and locally\cite{ravi2022degenerative}. Additional techniques are used to synthesize images that are high resolution, more individualized, and consistent with atrophy patterns associated to AD. Unlike VAEs which typically impose a Gaussian distribution on the latent space through their encoder, GAN models typically offer a uniform distribution where no point is more likely than any other, and cannot be directly used for statistical modeling to classify trajectories as normal or abnormal. 

\subsection{Our Contribution}
In this work, we aim to develop an approach that combines the complementary strengths of the two highlighted previous methods.  We maintain an interpretable latent space in the VAE framework, but introduce a new architecture to condition on past images leading to more realistic and higher resolution outputs. This is achieved by a novel double-encoder CVAE architecture to model changing pixel values. Our model takes in conditional inputs of two data types: a previous image and other demographic information (age, elapsed time and disease status), leading to a novel network architecture where the encoder and decoder don't exhibit the symmetry typical in VAE work. This approach allows prediction for an arbitrary elapsed time, i.e. multi-horizon prediction. 
Compared to previous VAE-based methods\cite{sauty2022progression}, we increase output resolution and subject specificity by incorporating a prior MRI. Compared to previous GAN-based approaches\cite{ravi2022degenerative}, our method produces a latent space, in which we can estimate a posterior probability of disease status for disease classification. 

Second, we introduce a novel training strategy to reduce computational requirement per sample and augment the dataset, where the network is trained on all possible pairs of images in a timeseries. The model we develop can predict future brain images with good accuracy, and we validate it in terms of mean square error (MSE) on the held out test and independent external dataset, while comparing to other baseline methods. Lastly, if a pair of scans has already been obtained, we show how our encoder can be used to build a likelihood ratio classifier. By focusing on individual trajectories in timeseries of MRI rather than population averages, our work has the potential to improve neurodegeneration biomarkers for AD.

We list our contribution as follows: i) we design a novel architecture that directly takes in both conditional images and demographic variables, ii) we propose a novel training strategy for computational cost reduction and data augmentation, iii) we propose a novel utilization of the latent space produced by our encoder for disease classification.

\section{Method}
\subsection{Data Preparation}
We obtain our dataset from the Alzheimer's Disease Neuroimaging Initiative (ADNI) database (\url{adni.loni.usc.edu}), led by Principal Investigator Michael W. Weiner, MD. The primary goal of ADNI is to measure the progression of mild cognitive impairment (MCI) and early Alzheimer's disease.
We include subjects from ADNI 1, 2 and 3 studies as of July 11, 2023, by searching for all MPRAGE scans. Here, we introduce a novel data preparation strategy to solve two problems we face in medical imaging datasets: data scarcity and computational limits. We structure our dataset so that each sample contains two images from the same patient, age and disease status when the first image was taken, and the time difference between the two images. For subjects with only one image, a pair of the same image is included, and the time difference is 0. For subjects with more than 2 images, we include all combinations of pairs both forward (positive time difference) and backward (negative time difference) in time. Including a negative time difference may or may not have direct clinical utility, but doubles our sample size. For example, if a person has 3 images we take the pairs \{(1,1),(1,2),(1,3),(2,1),(2,2),(2,3),(3,1),(3,2),(3,3)\}. This training strategy converts serial inputs into pairs of images, effectively augmenting our dataset and reducing computational costs per sample during training. We note this may introduce a bias where patients with more scans are overrepresented, and alternative sampling schemes will be the subject of future research. To simulate real data, often lower quality than carefully curated public datasets, we performed minimal quality check and did not discard any images.

For scalar conditional variables, we standardize age and time difference to mean 0 and variance 1. ADNI groups patients into 6 categories: cognitive normal, subjective memory complaint, early mild cognitive impairment, mild cognitive impairment, late mild cognitive impairment and AD. We follow the same grouping and assign a number from 0 to 5, noting this is a reasonable approximation of slightly different disease categories across ADNI studies. We assume that this progression is ordered, so we use a single number rather than a one hot encoded representation. We take only the diagnosis for the first scan of the pair, assuming future diagnostic labels are unknown. For the training set, we constructed 15,579 pairs from 907 subjects. For the test set, we held out 4 subjects at random from each category, for a total of 24 subjects, 180 images and 430 pairs. Characteristics of our ADNI cohort are shown in Table \ref{tab:cohort_characteristics}. In addition to ADNI, we obtain an independent set from OASIS3\cite{lamontagne2019oasis} with 1085 subjects for external validation. We follow the same preprocessing procedure as above. Since OASIS and ADNI have different conventions for disease status, we keep consistency by grouping OASIS subjects into cognitive normal, cognitive impairment and AD, and assign them a number of 0, 3 and 5.

Each image was rigidly aligned to the 2020 MNI extended nonlinear symmetric average template \cite{fonov2011unbiased}. First, the template was aligned to each brain image using an affine transformation, 
accounting for contrast differences using code from \cite{tward2020diffeomorphic} \url{(https://github.com/twardlab/emlddmm)}. Second, a Procrustes method was used to project this transform onto the closest rigid transform\cite{challis1995procedure} by minimizing sum of square error in voxel locations. This ensured that size and shape differences would be modeled by our network, and not ``normalized away'' by registration. We found through visual inspection that  an \emph{accurate} affine transformation followed by projection led to less variability, relative to an \emph{inaccurate} rigid transformation directly.
Finally, we cropped and downsampled all images to $80\times80\times80$, 2mm$^3$ voxels due to computational limitations.

\begin{table}[!htb]
\centering
\caption{ADNI cohort characteristics stratified by disease status.}
\label{tab:cohort_characteristics}
\begin{tabular}{lcccccc}
    \hline
    \multicolumn{1}{c|}{}  & \textbf{Normal} & \textbf{SMC} & \textbf{EMCI} & \textbf{MCI} & \textbf{LMCI} & \textbf{AD} \\
    \multicolumn{1}{c|}{} &(n=219) &(n=73) &(n=253) &(n=86) &(n=140) &(n=136) \\
    \hline
    \multicolumn{1}{c|}{Male} & 104(47.5\%) & 34(46.6\%) & 133(52.6\%) & 53(61.6\%) & 71(50.7\%) & 71(52.2\%) \\
    \hline
    \multicolumn{1}{c|}{Age*} & 75.51 [6.91] & 71.87 [5.6] & 70.84 [7.23] & 76.27 [8.37] & 72.02 [7.92] & 74.92 [7.98] \\    
    \multicolumn{1}{c|}{\quad 50-70} & 49(22.4\%) & 35(47.9\%) & 123(48.6\%) & 21(24.4\%) & 54(38.6\%) & 29(21.3\%) \\
    \multicolumn{1}{c|}{\quad 70-80} & 112(51.1\%) & 33(45.2\%) & 99(39.1\%) & 33(38.4\%) & 63(45\%) & 75(55.1\%) \\
    \multicolumn{1}{c|}{\quad 80-90} & 56(25.6\%) & 4(5.5\%) & 31(12.3\%) & 32(37.2\%) & 22(15.7\%) & 31(22.8\%) \\
    \multicolumn{1}{c|}{\quad 90+} & 2(0.9\%) & 1(1.4\%) & 0 & 0 & 1(0.7\%) & 1(0.7\%) \\
    \hline
    \multicolumn{1}{c|}{Scans per} & 4.74 [2.76] & 2.44 [1.14] & 4.7 [1.87] & 6.53 [4.71] & 4.49 [1.68] & 3.53 [2.09] \\
    \multicolumn{1}{c|}{subject} \\
    \multicolumn{1}{c|}{Length (years)} & 2.48 [1.83] & 1.61 [0.95] & 2.48 [1.6] & 1.98 [1.44] & 2.24 [1.53] & 0.88 [0.69] \\
    \hline
    \multicolumn{5}{l}{*mean and [standard deviation] of age at first visit}
\end{tabular}
\end{table}

\subsection{Autoencoders, VAE and CVAE}
First, we give an overview of autoencoder models and their extensions. Autoencoders are a class of neural network methods that learn a low-dimensional representation of high-dimensional structured data\cite{ehrhardt2022autoencoders}. They consist of two parts: an encoder that projects high dimensional data into a latent space with lower dimensions, and a decoder that learns to map a point in the latent space back to its high dimensional representation. The latent distribution of an autoencoder is unknown, making inference difficult and prompting the need for VAEs\cite{kingma2013auto}. 

VAE models assume the samples $x$ are generated from a distribution conditioned on the latent space $z$, where $z\sim p_{\theta}(z)$ and $ x\sim p_{\theta}(x|z)$\cite{kingma2013auto}. In practice, posterior inference may be intractable and estimating the posterior distribution is challenging\cite{kingma2013auto}. To address this problem, a more tractable distribution, $q_{\phi}(z|x)$, is used to approximate the true posterior\cite{kingma2013auto}. Therefore, VAEs can be trained to efficiently sample from unknown distributions using the variational lower bound of the log-likelihood, which can be written as follows\cite{kingma2013auto}:
\begin{align}
    \mathcal{L}(\theta, \phi, x) = -D_{KL}(q_{\phi}(z|x)||p_{\theta}(z)) + E_{q_{\phi}(z|x)}[\log p_{\theta}(x|z)],
\end{align}
where $D_{KL}$ is the Kullback–Leibler divergence\cite{kullback1951kullback} that measures how similar two distributions are, and $E_{q_{\phi}(z|x^i)}$ measures the sum of squared reconstruction error where we assume $p_\theta(x|z)$ is Gaussian with mean predicted by our decoder and fixed variance 1. We use the common choice that $p_\theta(z)$ is multivariate standard normal, with no learnable parameters.

As with many medical applications, conditional variables offer additional information and can improve parameter estimation. A natural extension to the VAE is CVAE\cite{sohn2015learning}, which includes an additional variable $y$ as the conditional variable. The variational lower bound for CVAE to optimize is as follows\cite{sohn2015learning}:
\begin{align}
    \mathcal{L}_{\text{CVAE}}(\theta, \phi, x, y) = -D_{KL}(q_{\phi}(z|x,y)||p_{\theta}(z)) + E_{q_{\phi}(z|x,y)}[\log p_{\theta}(y|x,z)].
    \label{eq:loss}
\end{align}

\subsection{Our Double Encoder CVAE}
We present our novel model architecture, inspired by CVAE\cite{sohn2015learning}, to generate 3D MR images. While previous methods have conditioned on a learned representation of an image, our architecture allows for a direct conditional image input. In our model, the encoder and decoder are not ``symmetric'', but rather the encoder takes the form of a standard CNN with two image inputs, and the decoder takes the form of a U-net\cite{ronneberger2015u}. The latter allows the prior MR image and conditional variables, as well as the latent space representation to be used for decoding. 
A point in the latent space does not represent an image, but a transformation between images, linking VAE modeling and classical work with diffeomorphisms.

\subsubsection{Objective function} 
We follow the same underlying assumptions of the distribution process as CVAE\cite{sohn2015learning}, where MR images are generated from the distribution $p_{\theta}(x|z)$ conditioned on 4 variables: the base image, age, disease status and elapsed time between two images. Though we have multi-modality conditional variables, the generation inference remains the same. Therefore, our empirical objective function follows closely to that of CVAE\cite{sohn2015learning} in Equation \eqref{eq:loss}.

\begin{figure}[!htb]
\centering
\includegraphics[width=\textwidth]{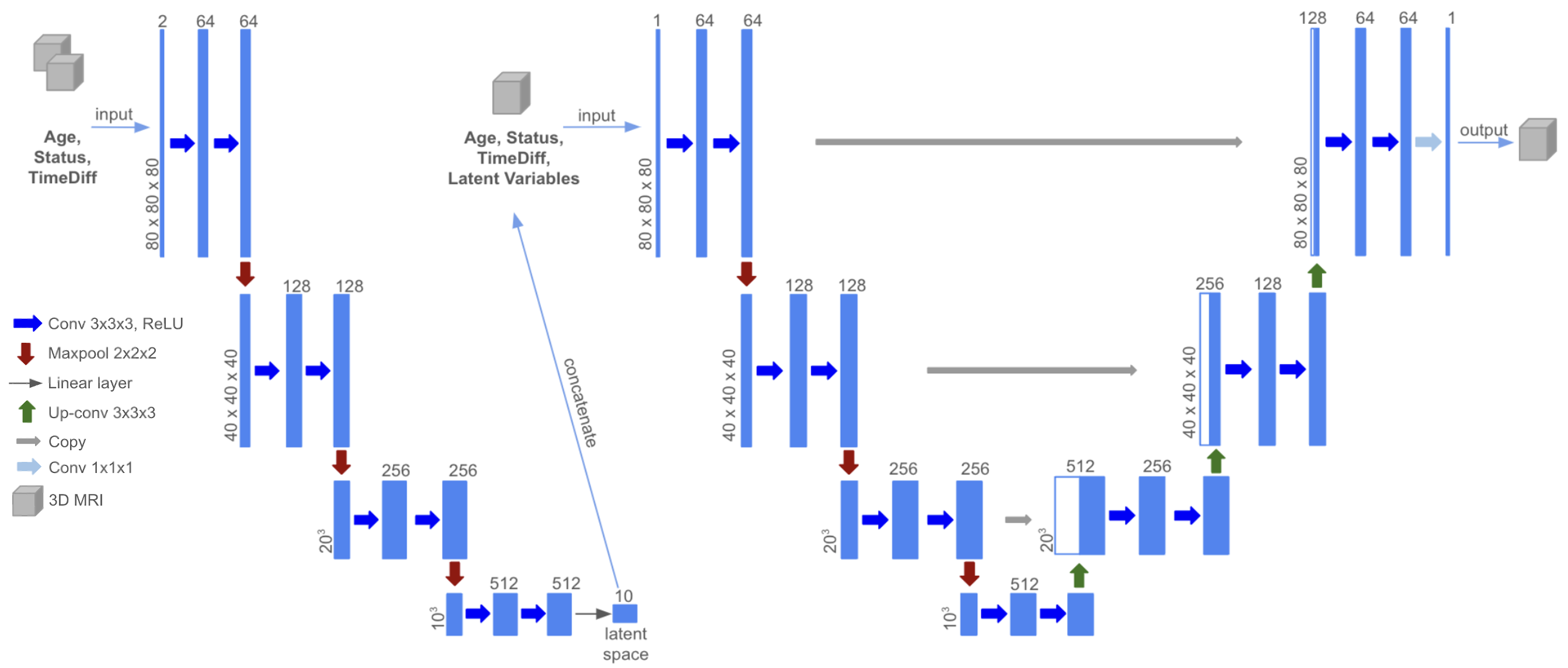}
\caption{Model architecture.} \label{fig1}
\end{figure}

\subsubsection{Model architecture} 
Due to our conditional variables' multi-modality nature, we implement a novel architecture to incorporate scalar variables (age, elapsed time, disease status) and a conditional 3D image. The overall architecture consists of an encoder and a U-net\cite{ronneberger2015u} shaped decoder. The encoder consists of 4 blocks, where each block contains 2 convolutional layers and 1 max pooling layer. We choose a kernel size of 3 and stride of 1 for all convolutional layers except the last, and a kernel size of 2 and stride of 2 for max pooling layers. Each convolution is followed by a group normalization layer\cite{wu2018group} with group size of 4 and a ReLU layer. During training, we take in an input of 2 stacked images and conditional variables, and project it down to a latent space of dimension 10 (i.e. we output 10 means and 10 log-variances). The decoder takes in the conditional image, latent space sample and scalar variables as inputs, and outputs the predicted image.
Since the downsampling branch of the U-Net can be thought of as an encoder, we call this architecture a ``double encoder CVAE''.
All non-image inputs are linearly transformed and added to each layer before the nonlinearity, following a strategy similar to ``timestep embedding'' common in modern diffusion models \cite{ho2020denoising}.
We show a visualization of our model in Fig.~\ref{fig1} and make our code available on GitHub at \url{https://github.com/rosie068/Double_Encoder_CVAE_AD}.

\subsubsection{Training procedure}
We used the Adam optimizer\cite{kingma2014adam} with learning rate 1e-5 and batch size of 4. We trained our model for 1000 epochs with early stopping on a single NVIDIA GeForce RTX 4090 GPU with a training time of 5 days.

\subsection{Comparison to alternative methods} \label{comp}
We attempted a comparison of our method to those described in section \ref{relate_work}, but were unable to produce satisfactory results using publicly available code and documentation.  We omit the comparisons, rather than casting these methods in a negative light. We suggest that a challenge style comparison, where authors can optimize parameter selection for their own methods and have them run automatically on a hidden test set, would be more appropriate for head to head comparisons.  In this work we compare to several simpler models to put our method into context and understand typical values of our figures of merit.  Our goal is to provide insight into the challenge of longitudinal image prediction by developing a new approach, not  claim superiority or inferiority. For a simple baseline, we use the conditional image as a prediction of the future image, and the other methods we compare to are described below.

\subsubsection{Low rank linear prediction} \label{lin}
For $N$ individuals, we take the first and last images in their trajectories and form two matrices $X$ and $Y$, both of size $80^3$ (number of pixels) by $N$. 
We include demographic variables by appending additional rows to $X$ for age, status, elapsed time and  1 (for mean). We perform singular value decomposition: $X = USV^T$, and take the first 10 (the same dimension as our CVAE latent space) or 100 (for better results) singular 
vectors as the latent space representation. To estimate the future MRI $\hat{Y}$ for individual $i$, we use the first few singular values (indicated by $ \tilde \cdot $) and calculate as: $\hat{Y_i} =  Y\tilde V \tilde S^{-1} \tilde U^TX_i$

\subsubsection{VAE with linear mixed effect estimation}
We train an autoencoder using the architecture described in \cite{sauty2022progression}, and fit a linear mixed effects (LME) model in the latent space using the statsmodels package in python \cite{seabold2010statsmodels}.  While other authors \cite{zhao2019variational} have noted that this approach should be improved upon, it provides a simple model based on standard VAEs for comparison. Images are predicted using the following procedure.  First a baseline image is passed through the encoder. Second, its latent space representation is shifted based on our LME model parameters, in a manner depending on time difference and disease status.  Third, the resulting vector is passed through the decoder. 

\section{Results}
\subsection{Trajectory prediction}
\begin{figure}[!htb]
\centering
\includegraphics[width=\textwidth]{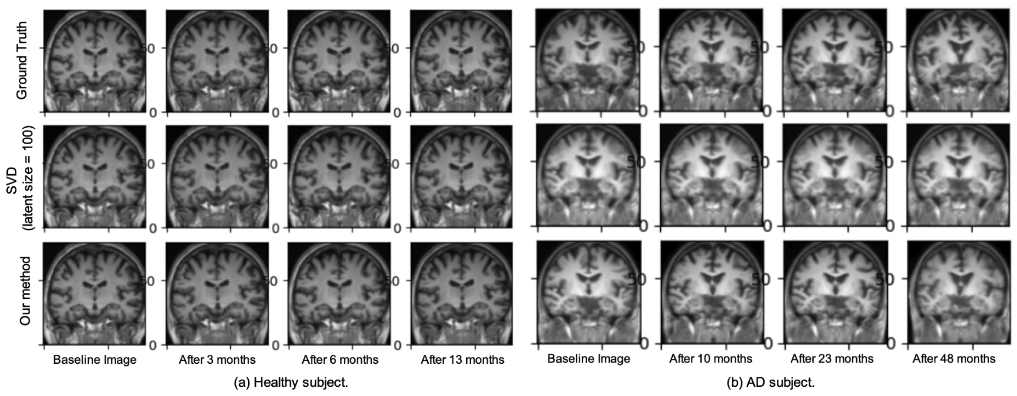}
\caption{Sample prediction  in the held-out test set.} \label{recon}
\end{figure}

First, we compare results for trajectory reconstruction in the held-out test set. In Fig. \ref{recon}, we visualize sample reconstructions for randomly selected healthy and AD individuals in the test set. As a healthy brain does not change much over 1 year, the ground truth does not have much variation and both the 100 dim. SVD method and our method perform well. In the diseased trajectory, we observe a noticeable enlargement in the ventricles, which is predicted well by our method but not the linear methods. The 10 dimensional SVD and VAE\plus LME model did not produce satisfactory images, and we do not show their outputs.

\begin{figure}[!htb]
\centering
\includegraphics[width=\textwidth]{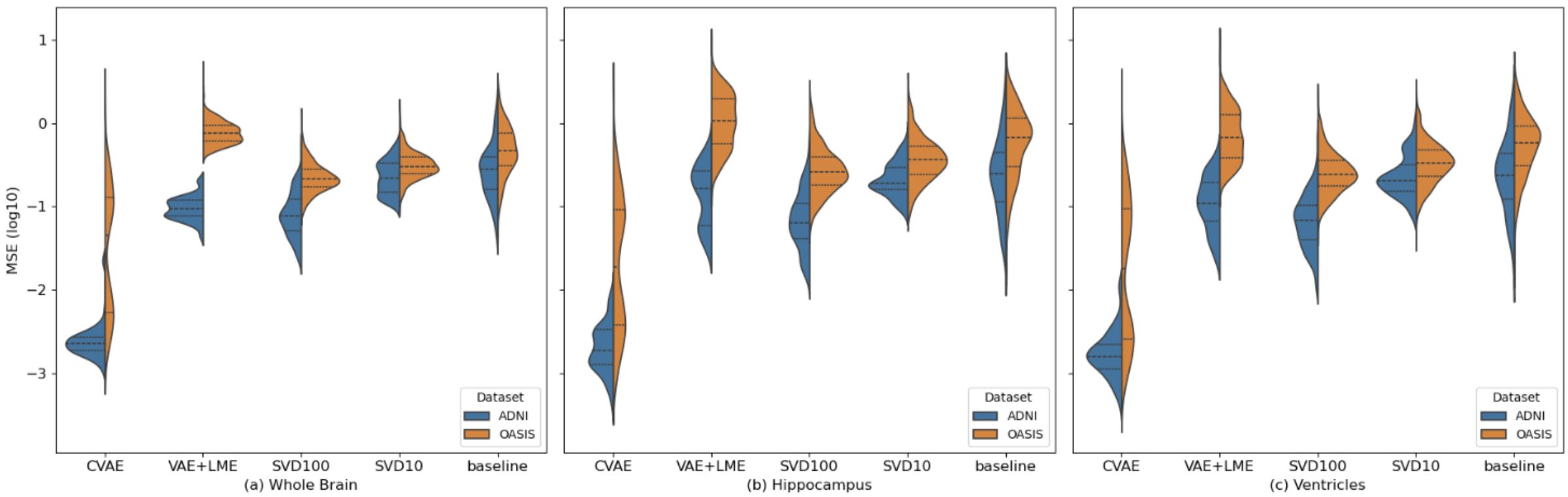}
\caption{MSE (log scale) in test (left) and external (right) validation set in 3 ROIs.} \label{MSE}
\end{figure}

For a quantitative comparison, we calculate mean squared error (MSE) in test and external validation datasets for 5 methods: our double encoder CVAE, VAE\plus LME model, SVD with both latent space sizes, and baseline. We conduct comparisons in three regions of interest (ROIs) to reflect global and local performance: the entire image, a region surrounding the hippocampus and a region surrounding the ventricles (both associated with AD progression\cite{evans2010volume,apostolova2012hippocampal}). ROIs were identified during image registration, using an affine transform of the MNI atlas and dilating annotations by several pixels. Out of 230 pairs, our model achieves the best performance in 202 pairs (88\%) in the whole brain and 218 pairs in both the hippocampus and ventricles (95\%). In the OASIS validation set, our model achieves the lowest MSE across all three regions, but the distribution has longer tails. Out of 1085 pairs of images, our model achieves the best performance in 855 pairs (79\%) in the whole brain, 889 pairs (82\%) in the hippocampus, and 922 pairs (85\%) in the ventricles. In Fig.~\ref{MSE}, we visualize the distribution of MSE on a log scale and show our proposed model reconstructs trajectories with more accuracy globally and locally. While still performing best, our method seems to follow a bimodal distribution with long tails in OASIS data, an observation which warrants further investigation.

\begin{figure}[!htb]
\centering
\includegraphics[width=\textwidth]{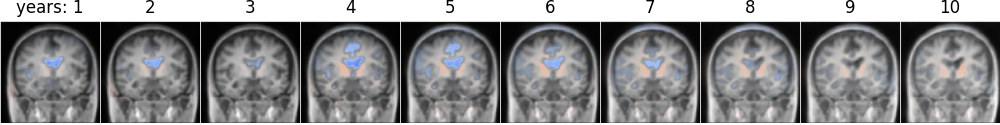}
\caption{Ten year prediction from one base image. Incremental changes (non-cumulative) shown via optical flow divergence, blue: expansion, red:  contraction.} \label{extra}
\end{figure}

In addition, we demonstrate our model's ability to interpolate between images and extrapolate beyond the last image in the timeseries. We generate images by sampling a standard multivariate normal in the latent space, and pass it through the decoder with the original image and scalar variables. The random latent sample is fixed across all panels, and only the elapsed time is changed. In Fig.~\ref{extra}, we predict the trajectory annually over 10 years for one individual in the test set with disease status 4 (LMCI) and only two real images available (extrapolation beyond the 4th year). We visualize structural changes over time by calculating a smooth optical flow estimate, and overlaying the divergence of the flow field.  Values close to 0 are transparent, positive (expansion) is shown in blue, and negative (contraction) is shown in red.  We observe  expansion in the ventricular space over time including its inferior horn in the temporal lobe, and contraction of nearby brain structures, in accordance with the AD aging process.

\subsection{Latent space disease status probability estimation}
\begin{figure}[!htb]
\centering
\includegraphics[width=0.33\textwidth,clip,trim=0in 5.4in 0in 0in]{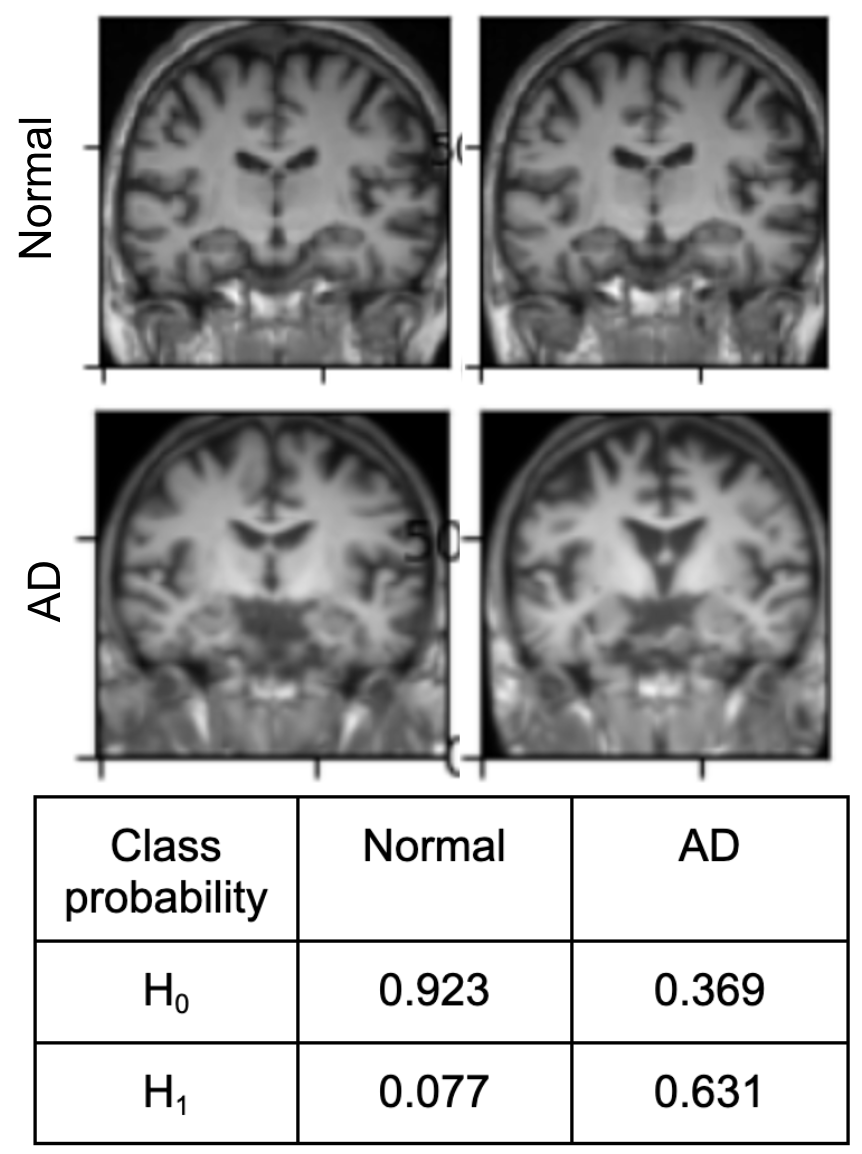}%
\includegraphics[width=0.33\textwidth,clip,trim=0in 2.7in 0in 2.7in]{hypothesis_test_table.png}%
\includegraphics[width=0.33\textwidth,clip,trim=0in 0.0in 0in 5.4in]{hypothesis_test_table.png}
\caption{Hypothesis testing for disease status, $H_0$: null, $H_1$: AD.} \label{hypo}
\end{figure}

In addition to trajectory estimation, we propose a practical use of the latent space to estimate a posterior probability of a subject's disease status given a pair of images. For subjects in the test set, we set the disease status to both normal and AD to obtain two sets of latent space representations. Let $\mu_{i,j}$ be the $j$th dimension of the encoder output when  disease status is $i$ (0 for healthy and 5 for AD), and define $f_i = \prod_{j=1}^{10}(2\pi)^{-1/2} \exp(-\mu_{ij}^2/2)$ (taking a product over the 10 dimensional outputs, reflecting our standard normal latent space distribution). Our posterior probability of disease status 0 is defined as $p_0 = f_0/(f_0+f_5)$. We show two examples of disease classification analysis in Fig.~\ref{hypo}, one healthy and one AD (same subjects as Fig.~\ref{recon}), where the posterior probability significantly favors the hypothesis corresponding to their true disease status.

\section{Discussion}
In this work, we present our double encoder CVAE, a novel architecture that predicts 3D MRI at an arbitrary time given a conditional prior image, age, disease status, and the future time. We compared our model on two datasets against other simpler methods, and showed our model is better at tracking both global and local changes during the aging process associated with AD. Our goal is to design a predictive model that operates on images and demographics only. While ADNI provides other biomarker information such as amyloid and tau pathology, we do not include them in this work but will consider in the future. One strength of our work is that our model learns aging patterns from populations without losing specificity when making individualized predictions. Given a baseline image, it could be used to understand potential trajectories of healthy aging or disease. Another strength is that our model produces a latent space for downstream inference, as demonstrated by our likelihood ratio classifier example. Our model has characteristics similar to ``image translation'' like the popular pix2pix model\cite{isola2017image}. However, pix2pix is deterministic\cite{raad2024conditional}, whereas our model allows users to explore uncertainty in future trajectories. Lastly, model predictions could be used for downstream analysis, including treatment effect comparison (where our model predicts the trajectory with no intervention).

One limitation of this work is that we were unable to perform head-to-head comparisons with state-of-the-art methods, and we suggest this could be addressed in the future in a challenge framework, where each author can run their own code. Another is that it is not straightforward to incorporate more than one previous scan when making predictions, unlike in \cite{sauty2022progression}. Future work will incorporate serial conditional images to increase predictive performance. Lastly, our test set size was relatively small due to a smaller cohort size compared to some computer vision datasets. 
With potential treatments for AD now approved, early diagnosis has become critical. Our model can generate synthetic datasets that can potentially be used for downstream tasks such as anomaly detection and classification. This work, which seeks to improve measures of neurodegeneration as a biomarker of AD by leveraging timeseries analysis, has the potential to impact treatment decisions.

\clearpage
\bibliography{paper-0010}
\end{document}